# Sub-Footprint Effect Correction in FW-LiDAR Point Clouds via Intra-Footprint Target Unmixing

Zhen Xiao, Yanfeng Gu, *Senior Member*, *IEEE*, and Xian Li, *Member*, *IEEE*

***Abstract*—Sub-footprint target mixing within a laser footprint significantly increases LiDAR intensity uncertainty, especially in complex environments where heterogeneous materials inside one footprint cause nonlinear distortions that impair intensity-based applications. However, the forward mixing inherent to the single-pixel detection mode of LiDAR systems blurs sub-footprint contributions, making sub-footprint effects difficult to address effectively in existing studies. To address this issue, we introduce a novel, physics-based framework that explicitly resolves sub-footprint intensity correction in full-waveform LiDAR (FW-LiDAR) point clouds. The key innovation is to make the otherwise implicit intra-footprint mixing process explicit: we first develop a spatiotemporal laser-beam distribution model to physically characterize within-footprint forward mixing of multi-target returns. Building on this formulation, we incorporate ancillary information including waveform parameters and surface geometry as constraints to pose a well-defined inverse unmixing problem and decompose each footprint into fractional contributions from multiple sub-targets. We then recover sub-footprint–corrected intensities by inverting the observed mixtures through a unified combination of parametric and model-driven approaches. To the best of our knowledge, few prior studies explicitly establish sub-footprint inversion and correction within a single laser footprint, and our framework offers a principled, physics-grounded solution. Experiments on both controlled and real-world LiDAR datasets demonstrate that the proposed method significantly enhances semantic separability across heterogeneous targets and intensity consistency across homogeneous targets.**

## I. INTRODUCTION

Light Detection and Ranging (LiDAR) has become a critical technology for acquiring high-resolution 3D information of Earth’s surface in a wide range of applications, including terrain classification [1], land cover analysis [2-3], and object recognition tasks [4]. In addition to spatial coordinates, LiDAR systems also capture the returned signal intensity, which reflects the target surface's material properties, geometry, and orientation [5]. This intensity information is increasingly used for urban mapping [6], vegetation analysis [7], digital elevation modeling [8], and data fusion [9-10] etc.

Manuscript received XX xx, 2025. This work was supported in part by the Major Scientific Instrument Development Program of the National Natural Science Foundation of China under Grant 62327803 and in part by the National Science Fund for Outstanding Young Scholars under Grant 62025107. (*Corresponding author: Xian Li.*)

The authors are with the School of Electronics and Information Engineering, Harbin Institute of Technology, Harbin 150001, China (e-mail: xianli@hit.edu.cn).

However, raw LiDAR intensities are influenced by multiple factors—including sensor settings, atmospheric attenuation, target reflectance, angle of incidence, and range—complicating their interpretation [11]. Prior studies therefore caution against using uncorrected intensities directly for feature construction or parameter retrieval, owing to uncertainties introduced by scan geometry [12–13]. By contrast, radiometrically corrected intensities that account for these effects provide more accurate and reliable three-dimensional (3D) object characterization [14–16]. Ideally, for a given natural surface, the intensity should be invariant with respect to range and angle of incidence. In practice, variations in surface orientation and sensor–target range produce substantial differences across angles and distances—collectively referred to as the LiDAR system’s radiometric (scan-geometry) effects.

Among these factors, the footprint-related sub-pixel mixing effect—referred to as the sub-footprint effect—is particularly pronounced in heterogeneous environments [17]. Due to the finite divergence of the laser beam, each emitted pulse illuminates a spatially extended footprint that may span multiple surface types with distinct reflectance properties. However, because LiDAR operates in an inherent single-pixel detection mode, the detailed interaction between a pulse and multiple within-footprint targets cannot be fully captured in space; instead, the system records only the backscattered photon energy in the order it arrives at the sensor, forming a waveform that effectively represents a collapse of spatiotemporal information into a one-dimensional signal. In this setting, the returned intensity is no longer representative of a single surface, but rather a nonlinear mixture of contributions from multiple intra-footprint targets. Moreover, the spatiotemporal collapse tends to bury sub-footprint effects within the waveform, making them difficult to explicitly express and correct, and leaving point-cloud intensities persistently contaminated by intra-footprint mixing.

Conventional intensity correction methods primarily rely on empirical models such as range normalization and incidence-angle compensation [11]. While these approaches can reduce geometry-induced distortions in homogeneous regions, they often fall short in complex terrain or across abrupt material transitions, where intra-footprint mixing dominates—resulting in boundary blurring, semantic ambiguity in classification, and degraded consistency in multi-temporal or multi-platform analyses. However, dedicated methods for sub-footprint effect correction remain largely absent, leaving point cloud intensities persistently affected by sub-footprint effect.

To address this challenge, we propose a novel, physics-grounded framework for sub-footprint effect correction in FW-LiDAR. We explicitly model the spatiotemporal distribution of laser energy within each footprint and decompose the observed intensity into sub-region contributions. By formulating the intensity correction as a constrained inverse problem and solving it using constrained unmixing techniques, our approach reconstructs a more faithful estimate of the true target reflectance, even under severe sub-footprint mixing.

The main contributions of this paper are summarized as follows:

(1) **A physics-interpretable framework for correcting sub-footprint effects.** We develop a physically interpretable intensity correction framework that reconstructs the footprint-scale laser energy distribution and decomposes the measured intensity into sub-footprint contributions, thereby improving semantic fidelity—especially near material boundaries and across heterogeneous terrain.

(2) **Physics-grounded sub-footprint modeling.** We present a physically grounded model that makes intra-footprint target mixing explicit, jointly accounting for laser-beam divergence and within-footprint reflectance heterogeneity to characterize the sub-footprint effect in FW-LiDAR intensity measurements.

(3) **Constrained inverse recovery with Intra-Footprint sparse unmixing.** We formulate sub-footprint correction as a sparsity-regularized constrained inverse problem and solve it via intra-footprint non-negative unmixing. Extensive experiments on synthetic and real airborne FW-LiDAR datasets demonstrate consistent improvements in intensity consistency, estimation accuracy, and semantic separability.

The remainder of this paper is organized as follows. Section II reviews the related work on LiDAR intensity correction, sub-footprint effects, and spectral unmixing models. Section III presents the proposed correction framework based on intra-footprint target decomposition and constrained inverse modeling. Section IV provides experimental results and comparative evaluations on both controlled and real-world LiDAR datasets. Finally, Section V concludes the paper and discusses potential directions for future research.

## II. RELATED WORK

LiDAR intensity correction has received increasing attention in recent years, especially with the growing demand for radiometrically consistent point clouds in high-level interpretation tasks. In this section, we review three major lines of related research: (1) LiDAR intensity correction techniques, (2) LiDAR model construction for sub-footprint mixing, and (3) target unmixing and inverse modeling approaches.

### *2.1 LiDAR Intensity Correction Techniques*

Most conventional LiDAR intensity correction methods focused on distance and incidence angle effects. These include: Distance normalization, typically modeled via inverse-square decay $I \propto 1/d^2$ [18-20]. Incidence angle correction, often based on Lambertian reflectance or empirical cosine models [21-23].

Notable works include the calibration framework by Höfle and Pfeifer [24], who analyzed the influence of sensor parameters and derived reflectance correction from physical models. Other studies, such as by Kaasalainen et al. [25], addressed radiometric normalization through reference targets and laboratory measurements. However, these methods often assume homogeneity within the laser footprint and fail to address the intra-footprint variability encountered in real-world mixed surfaces.

### *2.2 LiDAR Model Construction for Sub-footprint Mixing*

Sub-footprint mixing is a well-known problem in passive remote sensing (e.g., hyperspectral imaging), where a single pixel may represent a mixture of multiple materials. In active sensing systems such as LiDAR, the issue manifests differently due to: The spatial extent of the laser footprint (up to tens of centimeters to meters); Sloped or uneven terrain causing variable incidence angles; Structural discontinuities such as walls, vegetation edges, or roof boundaries.

While several studies have discussed multi-return waveform deconvolution to handle vertical mixtures [26-29], lateral mixing within a single footprint remains underexplored. Some recent efforts attempt to derive footprint-based bidirectional reflectance models [30-31], but few directly address the decomposition of observed intensity in the spatial domain.

### *2.3 Target Unmixing and Inverse Modeling Approaches*

Target unmixing has been extensively studied in the context of hyperspectral imagery, where observed spectra are expressed as convex combinations of endmember signatures [32-33]. Sparse unmixing, non-negative matrix factorization, and regularized least-squares have become common approaches.

In LiDAR, analogous formulations can be employed by treating observed intensity as a mixture of reflectance components weighted by the footprint energy distribution. Recent advances in physics-informed inverse modeling, such as point-based deconvolution [26] and radiative transfer inversion [35-36], offer new opportunities for physically interpretable correction models.

However, these methods are seldom applied to LiDAR intensity correction, and existing works lack a consistent treatment of the spatial energy kernel and non-uniform ground composition within the footprint. This motivates the need for a dedicated unmixing-based correction framework.

While existing intensity correction methods address basic geometric and radiometric distortions, they largely overlook the sub-footprint target mixing problem. According to the information we have at present, the correction method for the sub-footprint effect of single-band LiDAR is still a research blank. This paper bridges the gap between LiDAR radiometric correction and unmixing theory by proposing a constrained decomposition framework rooted in laser beam modeling and physically based inverse analysis.

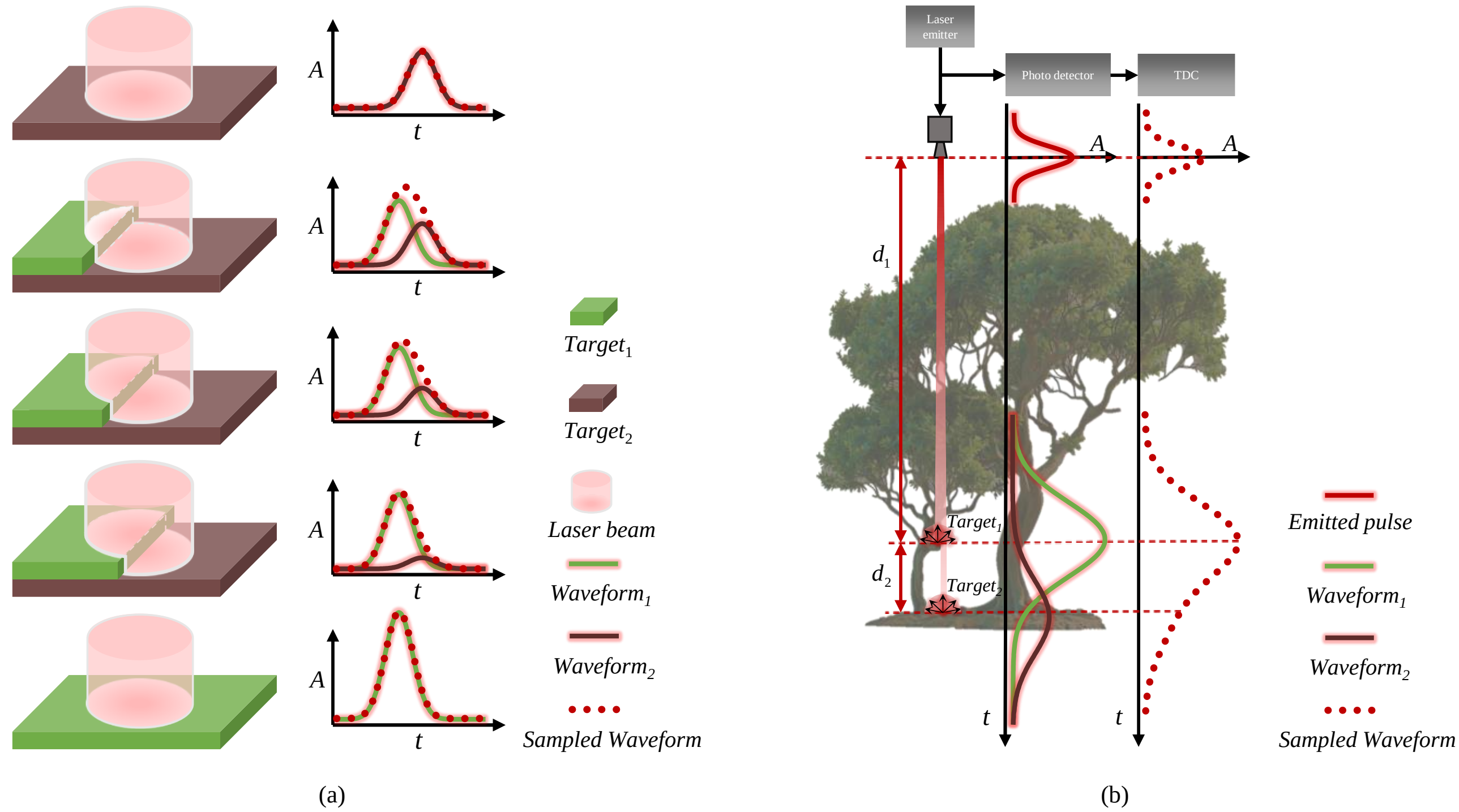

Fig. 1 Sub-footprint mixing effect. (a), Schematic diagram of the sub-footprint mixing effect. (b) The forward process of the sub-footprint effect.

## III. Method

This section presents the proposed FW-LiDAR sub-footprint intensity correction method based on intra-footprint target unmixing. The following text mainly introduces the three parts shown in Fig. 1(a), Fig. 1(b) and Fig. 2: (1) the sub-footprint mixing model, which describes how intra-footprint heterogeneity affects the observed FW-LiDAR intensity; (2) the construction of a physically based laser–target interaction forward model that incorporates footprint energy distribution and geometry; and (3) the formulation of a constrained inverse problem for intensity recovery through unmixing.

### *3.1 Sub-Footprint Mixing Model*

FW-LiDAR systems perform active remote sensing by emitting laser beams and recording the corresponding backscattered waveforms. Due to the inherent divergence angle $\beta$ of the emitted beam, the laser footprint expands with increasing range $d$, forming a circular footprint with an approximate diameter of $\beta d$ on the target surface.

In the ideal case, the laser beam is perpendicularly incident on a homogeneous and extended surface, such that the entire beam is intercepted by a single target, as illustrated in Fig. 1(a). However, real-world scenarios are often far more complex due to several contributing factors, including the scanning pattern of the FW-LiDAR sensor, the motion of the airborne or mobile platform, and the geometric and material heterogeneity of the scene. Additionally, the laser footprint may become obliquely projected onto sloped surfaces, partially intercepted by discontinuous surfaces, or distributed across multiple adjacent materials as shown in Fig. 1(a). In such heterogeneous surface conditions, a single laser footprint may simultaneously cover multiple materials (e.g., vegetation, asphalt, building edges), each contributing differently to the backscattered signal. Consequently, the recorded intensity at the footprint center does not reflect the properties of a single surface, but rather represents a weighted mixture of returns from all sub-regions within the footprint.

What further complicates this phenomenon is the fact that the spatial energy distribution within a laser footprint is typically non-uniform. Instead of being evenly spread, the emitted laser pulse energy is often well-approximated by a two-dimensional Gaussian distribution, with higher intensity concentrated near the center and gradually decreasing towards the edges. This anisotropic beam profile results in unequal energy illumination across the footprint area, meaning that targets located near the beam center contribute disproportionately more to the overall return.

The effective footprint diameter is usually defined based on the beam divergence angle $\beta$, measured at the $1/e^2$ power level of the Gaussian envelope. The $1/e^2$ contour of the Gaussian function is typically used to define the effective beam footprint boundary in radiometric FW-LiDAR models. This implies that the footprint boundary, and hence the spatial influence of a laser pulse, is inherently fuzzy and depends on both beam geometry and target topography.

We denote the footprint region $A$ as comprising $N$ sub-footprint areas $\{A_1, A_2, \ldots, A_N\}$, each with distinct surface reflectance $\rho_n$. Under this model, the observed intensity $I_{obs}$ is a mixture of weighted contributions from all sub-regions:

$$I_{obs} = \sum_{n=1}^{N} S_n \quad (1)$$

where $S_n$ is the partial return from region $A_n$, determined by the product of its energy exposure $w_n$, its surface reflectance $\rho_n$ and the geometric factors including range $d_n$ and incidence angle $\theta_n$. This formulation naturally leads to a linear mixing model, analogous to sub-pixel mixing in hyperspectral imagery, but adapted to the physical characteristics of FW-LiDAR energy distribution.

### 3.2 Forward model of laser–target interaction

In this section, we provide a detailed analysis of the backscattered waveforms generated by the interaction between the emitted laser beam and the target surface, including scenarios involving sub-footprint mixing effects. By explicitly incorporating the spatial energy distribution of the laser footprint and the heterogeneous composition of the illuminated surface, we aim to establish a physically grounded framework that captures the linear mixing behavior observed in practical FW-LiDAR data. This modeling is essential for accurately interpreting waveform distortions caused by partial beam interception, sloped surfaces, and multi-material returns within a single laser footprint.

**Vertically incident**

When the laser beam is vertically incident on the target surface, the energy distribution within the resulting footprint is typically well-approximated by a circularly symmetric two-dimensional Gaussian function. The irradiance $I(x,y)$ at any point $(x,y)$ on the footprint plane can be modeled as:

$$I(x,y) = I_0 \exp\left(-\left(x^2+y^2\right)/2\sigma^2\right) \quad (2)$$

where $I_0$ is the peak intensity at the footprint center, $\sigma$ is the standard deviation of the Gaussian profile, which is related to the beam divergence angle $\beta$ and the range $d$ via $\sigma \approx \beta d / 2\sqrt{2\ln 2}$, $(x,y)$ represents the local coordinates centered at the beam axis. This Gaussian approximation effectively captures the spatial decay of energy from the center to the edges of the footprint, and serves as a fundamental component in modeling intra-footprint energy-weighted mixing.

**Obliquely incident**

When the laser beam is obliquely incident on the target surface such as over sloped terrain, the originally circular Gaussian footprint undergoes both geometric deformation and spatial redistribution of energy. Specifically, the footprint shape changes from a circular disk to an approximate elliptical region due to the projection of the circular beam cross-section onto an inclined plane. The semi-major and semi-minor axes of the resulting ellipse depend on the incidence angle $\theta_n$ and the slope direction of the surface. The Gaussian energy distribution also becomes anisotropic in this case. The irradiance $I(x,y)$ is no longer circularly symmetric but elongated along the direction of projection. This can be modeled as:

$$I(x,y) = I_0 \exp\left(-[x \quad y]\Sigma^{-1}[x \quad y]'/2\right) \quad (3)$$

where $\Sigma = \left[\sigma_x \quad \sigma_y\right]'\left[\sigma_x \quad \sigma_y\right]$ is the covariance matrix defining the elliptical Gaussian footprint; $\sigma_x$ and $\sigma_y$ denote standard deviations along the principal directions of the projected footprint; $(x,y)$ are rotated coordinates aligned with the footprint axes. The Eq. (3) indicated that the targets located near the elongated axis may receive more energy than those on the shorter axis.

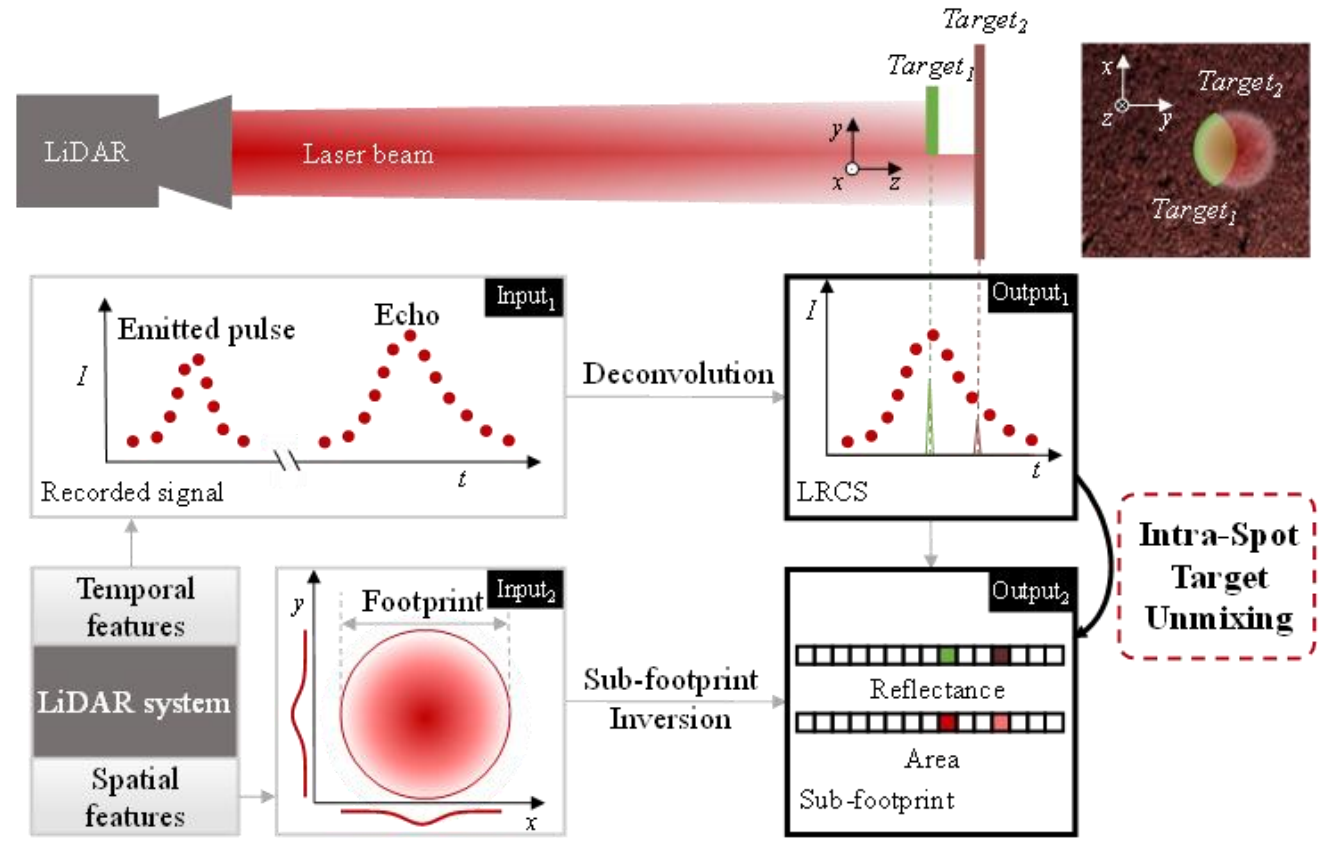

Fig. 2 Schematic of intensity correction for sub-footprint effects in FW-LiDAR.

**Sub-Footprint Intensity Mixing**

Let the footprint area $A$ be discretized into $N$ homogeneous sub-regions (e.g., different materials), each with a reflectance value $\rho_n \in [0,1]$, and occupying a spatial region $A_n \subset A$, such that:

$$A = \bigcup_{n=1}^{N} A_n, \quad A_i \cap A_j = \varnothing, i \neq j \quad (4)$$

Assuming the laser irradiance at point $(x,y) \in A$ is $I(x,y)$, and the total returned signal is the energy-weighted sum of local reflectance responses, the observed return intensity $R_{obs}$ is modeled as:

$$R_{obs} = \int_A \rho(x,y) \cdot I(x,y)\, dxdy \quad (5)$$

where $\rho(x,y)$ is the reflectance map of the surface under the footprint. If we approximate $\rho(x,y)$ as piecewise constant within each sub-region (i.e., $\rho(x,y) = \rho_n$ for

$(x,y)\in A_n$), then:

$$R_{obs}=\sum_{n=1}^{N}\rho_n\cdot\int_A I(x,y)\,dxdy=\sum_{n=1}^{N}\rho_n\cdot\omega_n \tag{6}$$

where $\omega_n=\int_A I(x,y)\,dxdy$ is the energy weight contributed by region $n$, and $\sum_{n=1}^{N}\omega_n=\sum_{n=1}^{N}\int_A I(x,y)\,dxdy=W_{total}$ is the total footprint energy. This is a convex combination of target reflectance, where $\omega_n$ depends on both the spatial location of material and the laser beam profile $I(x,y)$.

In the case of oblique incidence, the laser footprint projected onto the target surface becomes elliptical, and the irradiance density is modulated by the cosine of the incidence angle, i.e., $\cos(\theta_n)$. However, when the footprint is geometrically projected back to a plane normal to the beam direction, the apparent footprint area is reduced by a factor of $\cos(\theta_n)$, which precisely cancels the cosine attenuation of the irradiance. Therefore, under this geometric normalization, the footprint energy density remains consistent with that of the vertical incidence case, allowing the proposed sub-footprint mixing model to be directly extended to obliquely incident scenarios without additional correction factors. This geometric equivalence ensures that the spatial mixing weights $\omega_n$ in the unmixing model remain valid for sloped surfaces, as long as the energy distribution is computed on the beam-normal projection plane.

To enable practical numerical computation, we discretize the footprint into $M$ small pixels and define:

$\mathbf{i}\in\mathbb{R}^M$ : Laser energy vector at each pixel $i_j=I(x_j,y_j)$;

$\boldsymbol{\rho}\in\mathbb{R}^M$ : Reflectance map $\rho_j=\rho(x_j,y_j)$;

$\mathbf{m}\in\mathbb{R}^N$ : Reflectance of $N$ materials;

$\mathbf{S}\in\{0,1\}^{M\times N}$ : Segmentation matrix, where $S_{jn}=1$ if pixel $j\in A_n$, else 0.

Then the total observed intensity is:

$$R_{obs}=\mathbf{i}'\boldsymbol{\rho}=\mathbf{i}'\mathbf{S}\mathbf{m} \tag{7}$$

The energy weight vector is $\boldsymbol{\omega}=\mathbf{S}'\mathbf{i}$. This model explicitly links geometry (footprint shape), radiometry (beam energy profile), and semantics (material types). It shows that even small proportions of high-reflectivity materials near the footprint center can significantly bias the observed return. It provides the physical foundation for inverse modeling or unmixing algorithms aiming to recover $\{\rho_n\}$ or semantic labels from mixed observations.

We do not explicitly incorporate the effect of laser range (i.e., the distance between the sensor and the target) into the sub-footprint mixing model. This decision is grounded in both theoretical justification and practical precedent. It is well established in FW-LiDAR radiometric modeling that the received intensity decays approximately with the square of the range due to geometric spreading and atmospheric attenuation. This range-dependent attenuation is largely systematic and can be corrected using well-documented radiometric normalization models, or proprietary sensor-based range calibration protocols. Therefore, in our framework, the range-related intensity decay is treated as a pre-processing step and independently removed before sub-footprint unmixing.

When waveform data are available, the backscattered signal is captured as a high-resolution temporal curve, representing the energy contributions from different range bins within the laser footprint. This motivates the extension of our sub-footprint intensity mixing model into a spatiotemporal domain. We generalize the energy distribution and reflectance fields as time-dependent functions:

$$I(x,y)\rightarrow\mathbf{I}(x,y,t),\ \rho(x,y)\rightarrow\boldsymbol{\rho}(x,y,t) \tag{8}$$

This formulation allows the laser footprint to be interpreted as a 3D spatiotemporal volume, where each voxel contributes differently to the received waveform depending on its spatial position and depth (range). The temporal waveform profile encapsulates two distinct aspects of the laser–target interaction: Spatial characteristics of the laser footprint, corresponding to the intra-footprint energy distribution and material heterogeneity; temporal characteristics of the emitted and received pulses, including waveform stretching due to multiple path returns and scattering delays.

In our previous work, we conducted a detailed physical modeling of the FW-LiDAR waveform, explicitly accounting for both the spatial divergence of the beam and the temporal convolution introduced by system and environmental effects. The spatial divergence directly corresponds to the sub-footprint mixing phenomenon addressed in this study. Details of the waveform modeling and deconvolution methodology can be found in [26]. To isolate the intrinsic scattering response of the target, we employed a temporal deconvolution algorithm to remove the system-induced broadening effects. This procedure yields an estimate of the Laser Radar Cross-Section (LRCS) function:

$$R_{obs}=R_{obs}(t)=LRCS(t) \tag{9}$$

where $t$ is the temporal sampling index in waveform. Each LRCS value at time $t$ represents the backscatter strength of a specific depth slice within the footprint. This treatment effectively decouples the temporal blurring artifacts from the spatial mixing model, enabling a more accurate inversion of the intra-footprint composition based solely on the physical reflectivity of sub-targets.

Given the observed spatiotemporal intensity $R_{obs}(t)$, and the known or estimated laser energy distribution $\mathbf{I}(x,y,t)$, the goal is to recover the reflectance distribution $\boldsymbol{\rho}(x,y,t)$ across the sub-footprint volume. Thus, the above equation

becomes

$$LRCS(t)=\sum_x\sum_y diag\left(\mathbf{I}(x,y,t)\boldsymbol{\rho}'(x,y,t)\right) \tag{10}$$

where *diag* denotes the off-diagonal elements. This constitutes an inverse problem, where the observed signal is a nonlinear mixture governed by spatial energy weights.

### *3.3 Inverse Problem for Sub-Footprint Effect Correction*

For space targets, we assume that all targets at the same range share identical reflectance. This assumption is reasonable because the laser footprint is very small. When the laser illuminates objects at an identical distance, we treat them as a single effective target. In other words, within a given range bin along the distance dimension, only one target type is considered, and the reflectance is therefore defined as $\rho_t=\rho(x,y,t)$. We further assume that the entire beam footprint illuminates valid targets. Within the footprint, the illuminated area follows the beam's energy-density distribution, whereas the energy outside the footprint is zero. This assumption is justified by the laser's penetrability: along the beam path, targets will be irradiated until the footprint energy is effectively intercepted by the target. Except for an ideal blackbody or a target at infinite range, this assumption holds. For any spatial point, the normalized area density is denoted as $s(x,y,t)$. Over the entire time dimension, the target reflectance and normalized density-distributed area are expressed as $\boldsymbol{\Lambda}$ and $\mathbf{A}$, where $\boldsymbol{\Lambda}[t]=\rho_t, \mathbf{A}[t]=A_t, t=1,2,\cdots,T$. For voxel slice $t_i$ with no target distribution, its reflectance $\rho_{t_i}A_{t_i}=0$.

Considering the following constraints:

- For a single target under tilted incidence, they are distributed within multiple consecutive distance dimension slices, therefore the reflectance across consecutive slices should remain equal.
- The non-uniform energy density distribution of the laser beam requires the total target area to equal the beam footprint area.

Since both the reflectance of the material and the sub-footprint area are unknown, we construct the reflectance basis function $\boldsymbol{\Psi}\in\mathbb{R}^{M_1}$ and the energy density area basis function $\mathbf{S}\in\mathbb{R}^{M_2}$, where $\boldsymbol{\Psi}[m]=\psi_m, m=1,2,\cdots,M_1$ denotes the reflectance basis vector, and $\mathbf{S}[m]=s_m, m=1,2,\cdots,M_2$ represents the energy density basis vector. The target reflectance $\boldsymbol{\rho}(t)$ and effective area $\mathbf{A}(t)$ can be expressed as $\boldsymbol{\Lambda}=\mathbf{P}_1\boldsymbol{\Psi}$ and $\mathbf{A}=\mathbf{P}_2\mathbf{S}$. The $\mathbf{P}_1\in\{0,1\}^{T\times M_1}$ denotes the target reflectance representation matrix based on the reflectance basis vector. Similarly, $\mathbf{P}_2\in\mathbb{R}^{T\times M_2}$ represents the target effective area representation matrix based on the energy density basis vector, with elements $p_{ij}\geq 0$.

The target reflectance constraints—uniform reflectance under oblique incidence (identical reflectance across all points of a single target) and single-target exclusivity (each range-dimension voxel slice contains only one target)—yield the sparsity constraint $c_1=\left\|\mathbf{1}_{1\times T}\mathbf{P}_1\right\|_0$ and residual condition in the mathematical model.

$$\min_{\mathbf{P}_1\in\{0,1\}^{T\times M_1}} c_1 \quad \text{s.t.} \quad \mathbf{P}_1\mathbf{1}_{M_1\times 1}=\mathbf{1}_{T\times 1} \tag{11}$$

For the target effective area constraints (sparse target distribution along the range dimension and total effective area of all targets equaling the beam footprint area), the sparsity constraint $c_2=\left\|\mathbf{P}_2\mathbf{1}_{M_2\times 1}\right\|_0$ and residual condition can be derived.

$$\min_{\mathbf{P}_2>0} c_2 \quad \text{s.t.} \quad \mathbf{1}_{1\times T}\mathbf{P}_2=\omega[2k-1]_{k=1}^{M_2} \tag{12}$$

where $\omega\in\mathbb{R}^1$ is a parameter related to the voxel size and the energy density basis vector. The residual term of the LRCS is then formulated as $res=\left\|diag\left(\mathbf{P}_1\boldsymbol{\Psi}(\mathbf{P}_2\mathbf{S})'\right)-LRCS\right\|_2^2$

$$\begin{gathered}\min_{\mathbf{P}_1\in\{0,1\}^{T\times M_1},\mathbf{P}_2\geq 0} res+\lambda_1 c_1+\lambda_2 c_2 \\ \text{s.t.} \quad \mathbf{P}_1\mathbf{1}_{M_1\times 1}=\mathbf{1}_{T\times 1}, \mathbf{1}_{1\times T}\mathbf{P}_2=\omega[2k-1]_{k=1}^{M_2}\end{gathered} \tag{13}$$

where $\lambda_1$ and $\lambda_2$ denotes hyper parameters. After solving

TABLE I
TARGET LAYOUTS INFORMATION

| Index | Value |
|---|---|
| Material* | Leaf, PVC*, Ceramic, Aluminum, Soil, Concrete |
| Target number | 2,3 |
| Target interval | 60cm |
| Incident angle | 0 °, 10 °, 20 °, 30 °, 40 ° |
| Edge shape | Straight and angular shape, Arc shape |

*PVC represents Poly Vinyl Chloride.
*Incident angle: the angle between the laser beam and the first target's surface normal

TABLE II
CUSTOM FW-LIDAR SYSTEM INFORMATION

| Index | Value |
|---|---|
| Transmit pulse width | 4ns |
| Sampling Rate | 5GHz |
| Laser beam divergence | 0.3mrad |

TABLE III
SCANNED SCENES INFORMATION

| Picture and scene | Picture and description |
|---|---|
| 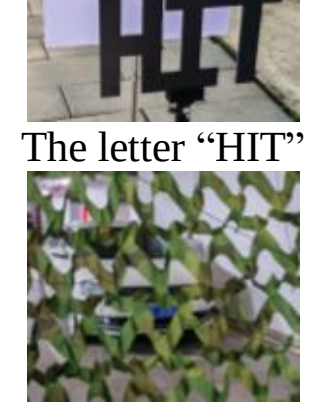 The letter "HIT" | Two PVC foam boards with 60 cm apart. The front board formed the black letters "HIT". |
| 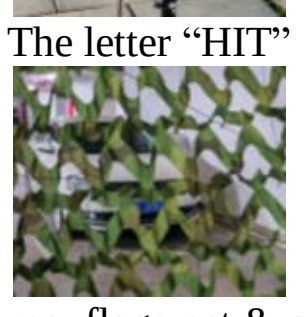 Camouflage net & car | A camouflage net was positioned approximately 2m in front of the car. |
| 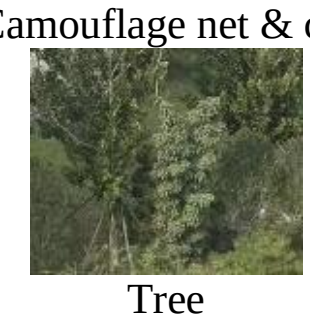 Tree | A foliage-rich tree with a complex three-dimensional structure. |

TABLE IV
RIEGL VZ2000 FW- LIDAR INFORMATION

| Index | Value |
|---|---|
| Transmit pulse width | 4ns |
| Sampling Rate | 500MHz |
| Laser beam divergence | 0.3mrad |

TABLE V
SHAPE COMPARISON BETWEEN SYNTHETIC AND REAL WAVEFORMS

| Target | 0° | 10° | 20° | 30° | 40° |
|---|---|---|---|---|---|
| Lf & Lf | | | | | |
| Lf & Lf & Sl | | | | | |

*SW denotes the synthetic waveform, and RW denotes the real waveform.

TABLE VI
SKEWNESS COMPARISON BETWEEN SYNTHETIC AND MEASURED WAVEFORMS

| Target | Value (Synthetic waveform skewness \| Real waveform skewness) | | | | | | | | | |
|---|---|---|---|---|---|---|---|---|---|---|
| | Straight and angular shape | | | | | Arc shape | | | | |
| | 0 ° | 10 ° | 20 ° | 30 ° | 40 ° | 0 ° | 10 ° | 20 ° | 30 ° | 40 ° |
| Lf & Lf | 1.31\|1.41 | 1.32\|1.43 | 1.32\|1.44 | 1.33\|1.45 | 1.32\|1.45 | 1.31\|1.30 | 1.32\|1.32 | 1.32\|1.37 | 1.33\|1.40 | 1.32\|1.38 |
| PVC & PVC | 1.31\|1.18 | 1.33\|1.23 | 1.43\|1.23 | 1.40\|1.24 | 1.41\|1.25 | 1.31\|1.35 | 1.33\|1.39 | 1.43\|1.43 | 1.40\|1.41 | 1.41\|1.49 |
| Al & Al | 1.19\|0.74 | 1.22\|1.18 | 1.28\|1.22 | 1.28\|1.26 | 1.32\|1.32 | 1.19\|1.20 | 1.22\|1.29 | 1.28\|1.30 | 1.28\|1.28 | 1.32\|1.29 |
| Cm & Cm | 1.28\|1.40 | 1.30\|1.34 | 1.37\|1.37 | 1.29\|1.32 | 1.31\|1.44 | 1.28\|1.25 | 1.27\|1.36 | 1.37\|1.22 | 1.29\|1.28 | 1.40\|1.28 |
| Lf & Lf & Sl | 1.06\|1.05 | 1.09\|1.04 | 1.11\|1.19 | 1.11\|1.11 | 1.12\|1.17 | 1.06\|1.05 | 1.09\|1.04 | 1.11\|1.19 | 1.11\|1.11 | 1.12\|1.17 |
| PVC & PVC & Cc | 1.07\|1.04 | 1.14\|1.18 | 1.26\|1.17 | 1.24\|1.07 | 1.25\|1.16 | 1.07\|1.04 | 1.14\|1.18 | 1.26\|1.17 | 1.24\|1.07 | 1.25\|1.16 |
| Al & Al & Cc | 0.97\|1.15 | 1.04\|1.16 | 1.13\|1.31 | 1.15\|1.38 | 1.17\|1.36 | 0.97\|1.15 | 1.04\|1.16 | 1.13\|1.31 | 1.15\|1.38 | 1.17\|1.36 |
| Cm & Cm & Cc | 1.03\|0.71 | 1.07\|0.66 | 1.15\|1.10 | 1.05\|1.07 | 1.08\|1.12 | 1.03\|0.71 | 1.07\|0.66 | 1.15\|1.10 | 1.05\|1.07 | 1.08\|1.12 |

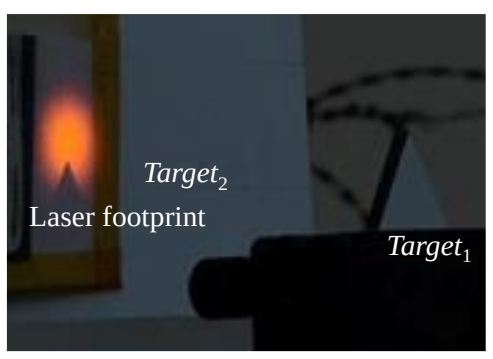

Fig. 3 An infrared-sensitive plate shows the sub-footprint effect.

for the effective area representation matrix $\mathbf{P}_2$, the distribution of targets within the beam footprint can be calculated, thus the intensity distortion caused by the distribution of sub-footprint can be corrected.

## IV. EXPERIMENT RESULTS AND ANALYSIS

Using challenging datasets collected with a small-footprint FW-LiDAR, we conducted fixed-position tests (controlled datasets) and scanning-imaging experiments (real-world datasets) to evaluate the proposed method. We first detail the data-acquisition setup and protocol, then test the suppression of sub-footprint effects and benchmark it against mainstream intensity-correction approaches, confirming the effectiveness of our method. Finally, we perform comparison studies to examine how different factors influence point-cloud intensity.

### *4.1 Data Description and Reprocessing*

To evaluate the effectiveness of the proposed sub-footprint intensity correction method based on intra-footprint target unmixing, we conducted both controlled synthetic simulations and experiments on real-world FW-LiDAR datasets.

**1) Controlled Scene Construction**

To obtain controllable data exhibiting sub-footprint effects, we built a fixed-beam FW-LiDAR system and configured a rigorously calibrated test field. Targets with different spacings, counts, incidence angles, and materials were arranged to emulate diverse real-world scenarios. The target layouts and key instrument specifications are summarized in Tables I, II. For brevity, we abbreviate selected material names and use these abbreviations in the subsequent table and figures: Leaf (Lf), Ceramic (Cm), Aluminum (Al), Soil (Sl), Concrete (Cc).

**2) Real-World Dataset**

To assess the impact of sub-footprint effects during practical scanning and to validate the proposed method, we performed scanning with a RIEGL VZ-2000 FW-LiDAR system. The scanned scenes and key instrument specifications are listed in Tables III, IV.

**3) System Response Removal**

In our work, we focus on mitigating sub-footprint effects; therefore, the raw waveform data were preprocessed to facilitate their observation and to evaluate the effectiveness of the proposed method. The preprocessing primarily involved removal of the FW-LiDAR system response. Because the receiver convolves the echo with the system impulse response [37-38]—which not only alters the waveform shape but also interacts with the dynamic gain circuitry—we performed extended-target calibrations for both systems at multiple ranges and incidence angles, and then applied a deconvolution algorithm to eliminate the system response.

TABLE VII
INTENSITY CORRECTION RESULTS FOR VARIOUS MATERIALS UNDER DIFFERENT SITUATIONS

| Situation | Lf (mean ±std) | PVC (mean ±std) | Al (mean ±std) | Cm (mean ±std) | Sl (mean ±std) | Cc (mean ±std) |
|---|---|---|---|---|---|---|
| Situation 1 | Lf & Lf<br>93 ±1.11 | PVC & PVC<br>292 ±3.41 | Al & Al<br>309 ±0.74 | Cm & Cm<br>286 ±1.19 | Lf & Lf & Sl<br>83 ±2.92 | PVC & PVC & Cc<br>126 ±0.99 |
| Situation 2 | Lf & Lf & Sl<br>91 ±1.73 | PVC & PVC & Cc<br>295 ±1.83 | Al & Al & Cc<br>318 ±2.17 | Cm & Cm & Cc<br>277 ±2.64 | -- | Al & Al & Cc<br>126 ±0.99 |
| Situation 3 | -- | -- | -- | -- | -- | Cm & Cm & Cc<br>132 ±1.41 |
| All situations | 92 ±2.34 | 294 ±4.52 | 314 ±4.37 | 282 ±4.62 | 83 ±2.92 | 124 ±7.74 |

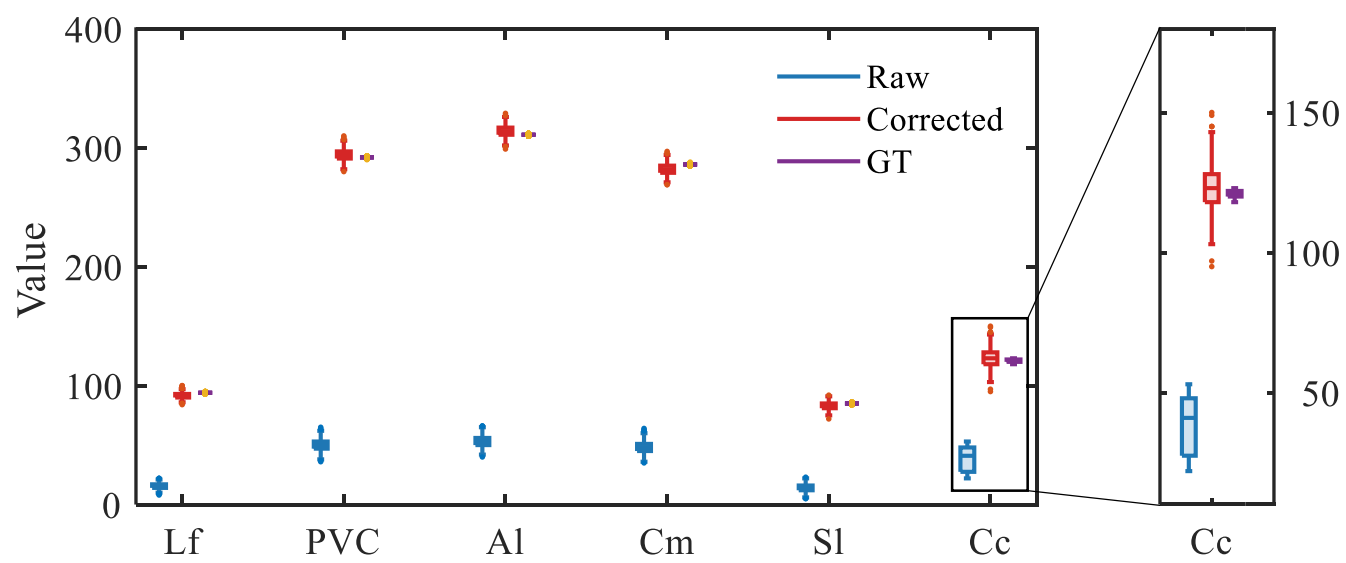

Fig .4 Comparison between the results and the ground truth before and after sub-footprint effect correction for targets of different materials.

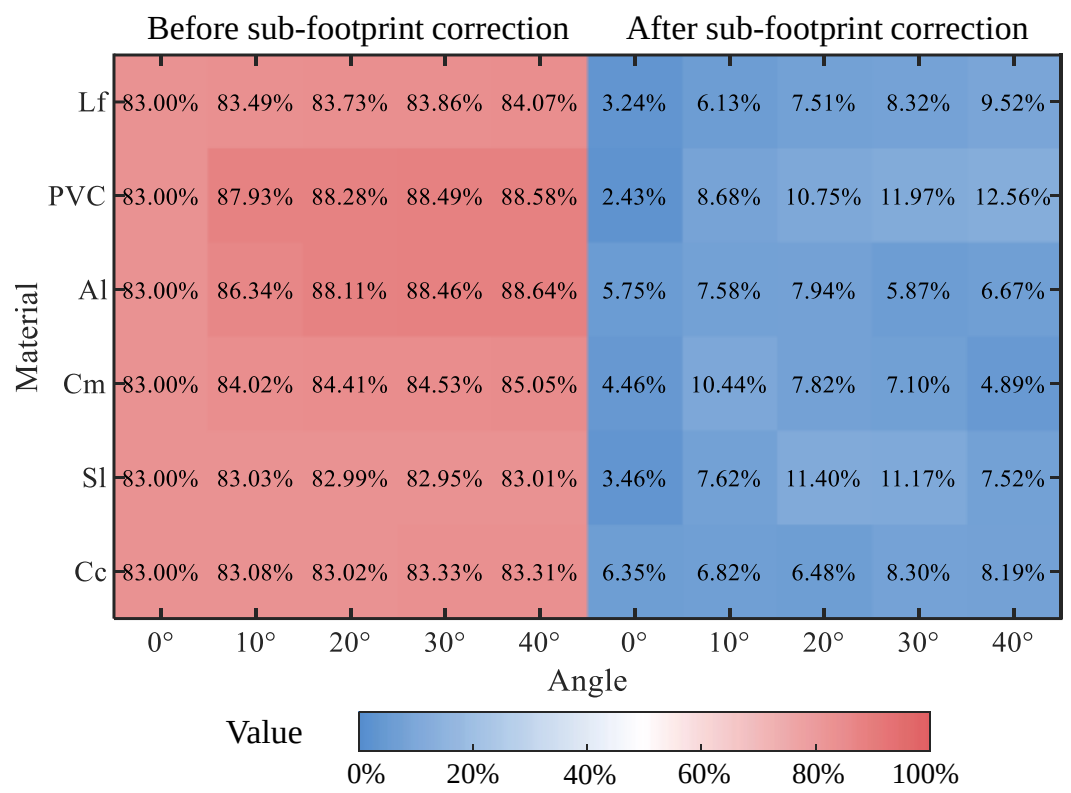

Fig .5. Relative errors before and after sub-footprint effect correction for targets of different materials at different angles.

### *4.2 Sub-footprint Effect Analysis*

To directly observe the sub-footprint effect, we used an IR-sensitive plate in the controlled data-acquisition experiments to record the sub-footprint of the laser footprint on the plate when the beam illuminated different targets as shown in Fig. 3. According to the forward model analyzed in the Methods, this sub-footprint effect further influences the amplitude and shape of the return waveform. Therefore, we synthesized multi-target data using waveforms free of sub-footprint effects—obtained when the laser footprint illuminated extended targets—and compared the amplitudes and shapes of these synthetic multi-target waveforms with those of multi-target waveforms that exhibit sub-footprint effects.

In Table V, we present a subset of waveforms corresponding to the “Straight” and “Angular-shaped” targets in the leaf two-target and three-target cases. As the waveform plots show, when sub-footprint effects are present, the amplitudes are smaller than those of the synthetic data, and as the incidence angle increases the waveform shape tends to skew toward one side.

To further quantify the differences between the synthetic multi-target waveforms and those with sub-footprint effects, we configured different scenarios for the various target materials listed in Table I, as shown in Table VI. The table reports, for each case, the skewness coefficient and amplitude for both the synthetic multi-target waveforms and the multi-target waveforms affected by sub-footprint effects. Comparing the entries shows that the latter generally have lower amplitudes and larger skewness coefficients. This is consistent with the predictions of the forward model in the Methods: in the presence of a sub-footprint effect, the target receives less laser energy and thus backscatters less energy; moreover, when there is a nonzero incidence angle, the sub-footprint further modifies the already nonuniform energy distribution within the laser footprint, thereby increasing the skewness coefficient.

From the above analysis, the existence of the sub-footprint effect is beyond doubt, and its impact on the intensity of the generated point-cloud data is unavoidable. It is therefore important to correct for the sub-footprint effect to obtain more accurate point-cloud intensities.

### *4.3 Sub-footprint Effect Correction*

The proposed method was tested on above challenging measured FW-LiDAR data, preprocessed by system response removal. Moreover, incidence-angle and range corrections were conducted by calibration data so that the improvement due independently to sub-footprint correction could be assessed.

We report both quantitative metrics and qualitative assessments to substantiate the improvements. The former, intended chiefly for controlled datasets from fixed-position scans, include within-class consistency (the stability of a material’s intensity across conditions) and agreement with a calibration reference (the concordance between sub-footprint–corrected intensities and calibration data). The latter, applied to scan-imaging datasets, primarily evaluate the effectiveness of sub-footprint correction via visualization of point-cloud intensities and histogram-based assessment of the tightening of same-material intensity distributions.

In the controlled-data experiments, we applied the proposed method to correct the sub-footprint effect and presented, for each material, the intensity-correction results under different scenarios in Table VII. From the results we see that, for the same material, the standard deviations indicate that, after intensity correction, the intensities are more consistent within the same scenario; comparisons of the per-scene means and the aggregated statistics across all scenes further show good consistency across scenarios.

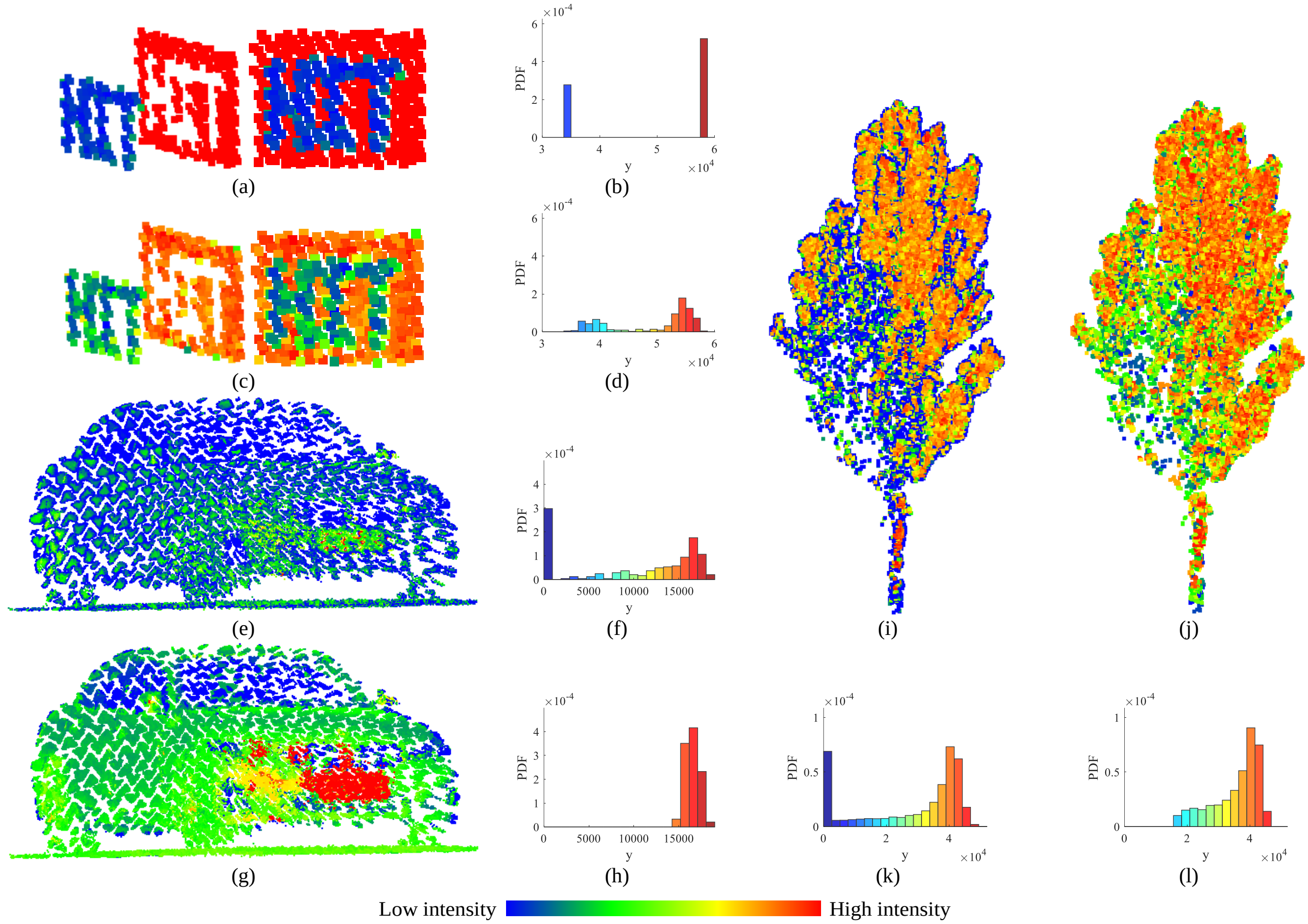

Fig. 6 Point cloud visualization and histogram of intensity distribution. (a)(b)(e)(f)(i)(k). Raw point cloud and histogram. (c)(d)(g)(h)(j)(l). Corrected point cloud and histogram.

To facilitate evaluation of the gains from sub-footprint correction, we compared, across all scenarios in Table VII, the corrected intensities for different materials with the raw data and with the ground-truth data (provided by waveforms obtained when the laser footprint illuminated extended targets), as shown in Fig. 4. The results indicate that, before sub-footprint effect correction, the intensities for different targets differ markedly from the ground truth, whereas after correction they are closer to the ground truth.

Regarding performance across different incidence angles, we compiled the sub-footprint effect correction results for each material, comparing them across scenarios and angles with the ground truth (Fig. 5). The analysis reveals that while the post-correction error increases with incidence angle, it still shows a clear improvement over the pre-correction data.

To provide a more intuitive comparison before and after correction, we used the scan data in Table III to render point-cloud intensities for visualization (Fig. 6), and we also plotted the intensity histograms for each point cloud. The patches in Fig. 6 (e) and (g) represent point clouds formed by the laser beam passing through the camouflage net and illuminating the car. From the visualizations, we observe pronounced sub-footprint effects along object edges: because the laser footprint illuminates multiple targets there, edge-contour points have noticeably lower intensities than points in extended-target regions. After sub-footprint correction, the edge-contour intensities become more consistent with those of same-material points in the extended regions. The intensity histograms likewise show a more concentrated distribution after correction, which is reasonable for relatively clean test scenes (observing only the test targets listed in Table III), and further corroborates the effectiveness of the sub-footprint correction.

### *4.5 Intensity Correction Comparison*

As no established methods exist for correcting the sub-footprint effect in single-wavelength FW-LiDAR, we did not include method-to-method comparisons. To further assess the gains in intensity correction, we used controlled datasets to partition the measured intensity into contributions from incidence angle, range, and the sub-footprint effect in Fig. 7.

In this study, the synthetic data were generated at an absolute range of about 50 m. For a relative separation of 60 cm, the induced change in intensity is negligible; therefore, the contribution of range appears small in Fig. 7. This does not deny the effect of range on intensity in general: when ranges differ substantially, the impact remains large. With respect to incidence angle, it shows that the intensity varies with angle in ways that are not consistent across materials; this relates to surface roughness, and targets cannot always be treated as ideal Lambertian surfaces. For smooth PVC and aluminum, the incidence angle is prominent, while for rougher materials like soil and concrete, the sub-footprint effect dominates. After separately correcting for incidence angle and for the sub-footprint effect, we quantified their respective contributions to intensity accuracy. Relative to the raw intensities, the angle correction increases accuracy by 1%–71% (excluding vertical incidence), whereas the sub-footprint correction increases it by 27%–95%, indicating that the sub-footprint effect plays a crucial role in determining intensity. Although in many extended-target scenarios (e.g., purely urban scenes) the sub-footprint effect is not common, it has a pronounced influence on edge delineation; in vegetation-rich scenes (e.g., the scanned tree), the sub-footprint effect is widespread.

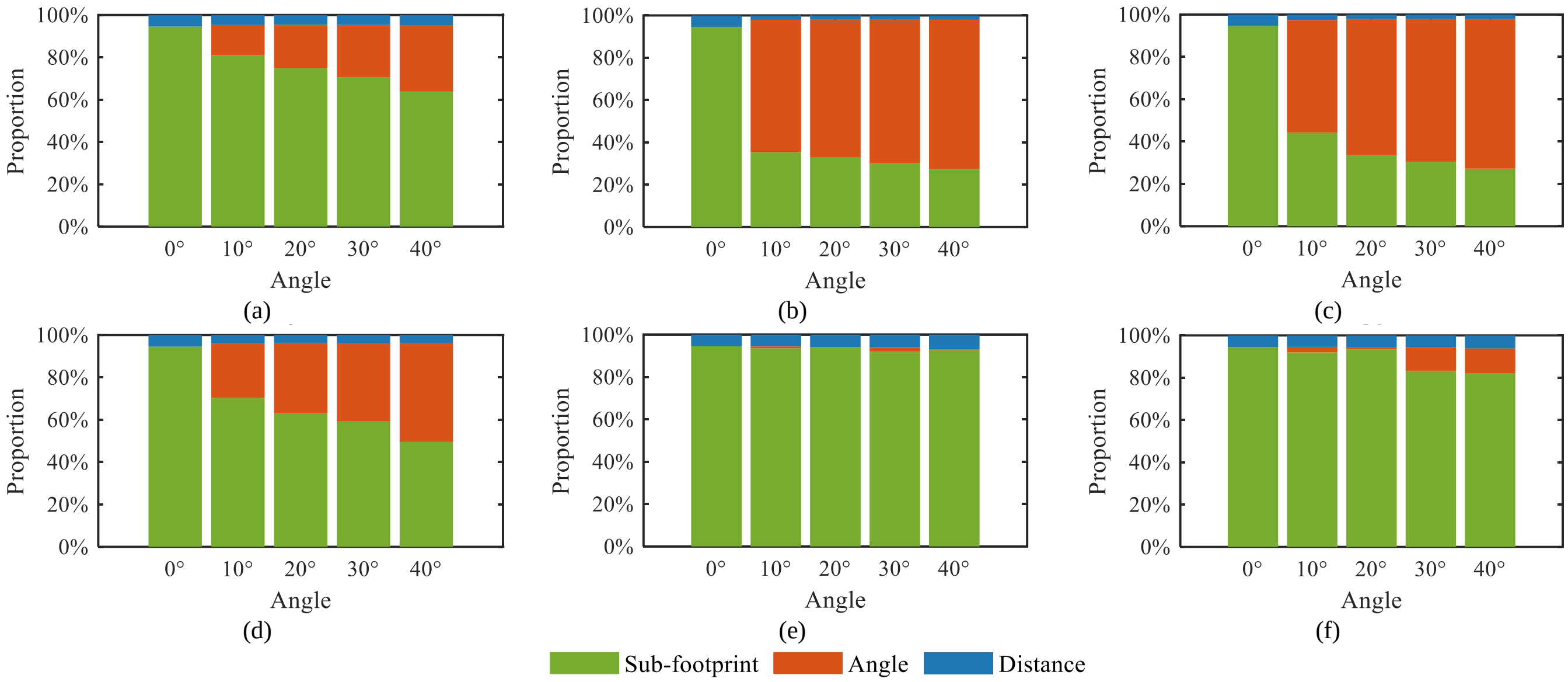

Fig. 7 Relative contributions of different factors to intensity. (a). Leaf. (b). PVC. (c). Aluminum. (d). Ceramic. (e). Soil. (f). Concrete.

## V. Conclusion

This paper presents a novel physically based framework for correcting sub-footprint intensity distortions in LiDAR point clouds by explicitly modeling and decomposing intra-footprint target mixtures. Unlike conventional approaches that rely solely on distance or incidence angle compensation, the proposed method reconstructs the laser footprint energy distribution and unmixes the observed return into fractional reflectance contributions from multiple ground targets.

By formulating the correction as a constrained inverse problem and leveraging non-negative unmixing techniques, we achieve accurate recovery of true reflectance values even in regions with strong spatial heterogeneity, such as object boundaries, sloped terrain, and composite surfaces. Experiments conducted on both synthetic scenes and real-world LiDAR data demonstrate substantial improvements in intensity consistency and semantic separability performance. In particular, our method reduced reflectance error by over 70%. Also, the method has limitation that it relies on accurate knowledge of the beam shape, footprint geometry, and does not consider the effect of multiple reflections. Overall, this work provides a new perspective on understanding and mitigating sub-footprint effects in LiDAR sensing and opens up new directions for high-fidelity intensity correction and physically interpretable remote sensing analysis. In future work, we aim to investigate deep learning methods that integrate physical models to streamline the sub-footprint unmixing process.

## Acknowledgment

This work was supported by the National Science Fund for Outstanding Young Scholars under Grant 62025107 and the Major Scientific Instrument Development Program of the National Natural Science Foundation of China under Grant 62327803.

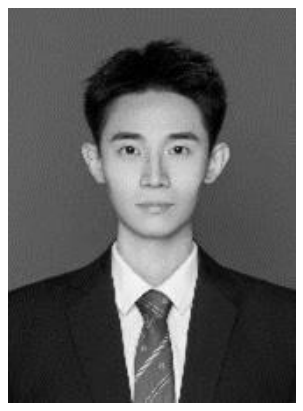

**Zhen Xiao** received the B.E. degree in electronics and information engineering from the Harbin Institute of Technology, Harbin, China, in 2020, where he is currently pursuing the Ph.D. degree in information and communication engineering. His research interests include UAV LiDAR point cloud processing and the application of point cloud in remote sensing.

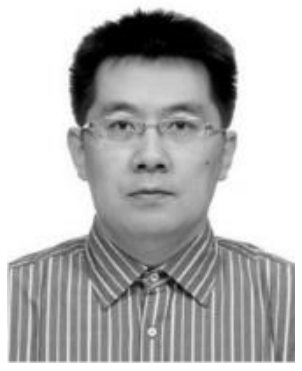

**Yanfeng Gu** (M'06-SM'16) received the Ph.D. degree in information and communication engineering from Harbin Institute of Technology, Harbin, China, in 2005. He joined as a Lecture with the School of Electronics and Information Engineering, Harbin Institute of Technology (HIT). He was appointed as Associate Professor at the same institute in 2006; meanwhile, he was enrolled in first Outstanding Young Teacher Training Program of HIT. From 2011 to 2012, he was a Visiting Scholar with the Department of Electrical Engineering and Computer Science, University of California, Berkeley, CA, USA. He is currently a Professor with the Department of Information Engineering, HIT, Harbin, China. He has published more than 100 peer-reviewed papers, four book chapters, and he is the inventor or coinventor of 20 patents. His research interests include space intelligent remote sensing and information processing, multimodal hyperspectral remote sensing, spaceborne time-series image processing.

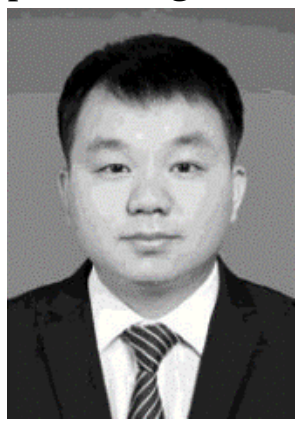

**Xian Li** (Member, IEEE) received the Ph.D. degree in instrument science and technology from the Harbin Institute of Technology (HIT), Harbin, China, in 2021. From 2018 to 2020, he was a Doctoral Researcher with the Department of Telecommunications and Information Processing, Ghent University, Ghent, Belgium, supported by the China Scholarship Council. He is currently an Assistant Professor with the School of Electronics and Information Engineering, HIT. His research interests include deep learning, hyperspectral remote sensing, and image analysis.